%% file: main.tex
\documentclass[10pt,twocolumn,letterpaper]{article}

\usepackage{cvpr}
\usepackage{times}
\usepackage{epsfig}
\usepackage{graphicx}
\usepackage{amsmath}
\usepackage{amssymb}
\usepackage{lipsum}

\usepackage{enumitem}

\usepackage{color}
\usepackage{xcolor,soul}
\sethlcolor{lightgray}
\usepackage{multicol}

\usepackage{pifont}
\usepackage{multirow}
\usepackage{rotating}
\usepackage{bm}

\usepackage[pagebackref=true,breaklinks=true,letterpaper=true,colorlinks,bookmarks=false]{hyperref}

\cvprfinalcopy 

\def\cvprPaperID{} 

\ifcvprfinal\fi
\begin{document}

\title{Knowledge-Based Visual Question Answering in Videos}

\author{Noa Garcia\\
Osaka University\\
\and
Mayu Otani\\
CyberAgent, Inc.\\
\and
Chenhui Chu\\
Osaka University\\
\and
Yuta Nakashima\\
Osaka University\\
}

\maketitle

\input{1-intro.tex}

\input{3-model.tex}
\input{4-evaluation.tex}
\input{5-conclusion.tex}

\section*{Acknowledgements}
This work was partly supported by JSPS KAKENHI No.~18H03264 and JST ACT-I.

{
\bibliographystyle{ieee_fullname}
\bibliography{egbib}
}

\end{document}

%% file: 1-intro.tex
\section{Introduction}
In standard visual question answering (VQA), answers are usually inferred by extracting the visual content of the images and generalising the information seen at training time. However, as the space of training question-image pairs is finite, the use of image content as the only source of information to predict answers presents two important limitations. First, image features only capture the spatial information of the picture, leaving temporal coherence in video unattended. Second, visual content by itself does not provide enough insights for answering questions that require reasoning beyond the image. To address these limitations, video question answering (VideoQA) \cite{tapaswi2016movieqa,lei2018tvqa} and knowledge-based visual question answering (KBVQA) \cite{wu2016ask,wang2018fvqa} have emerged as two independent fields.

\begin{figure}[t]
\centering
\includegraphics[width = 0.46\textwidth]{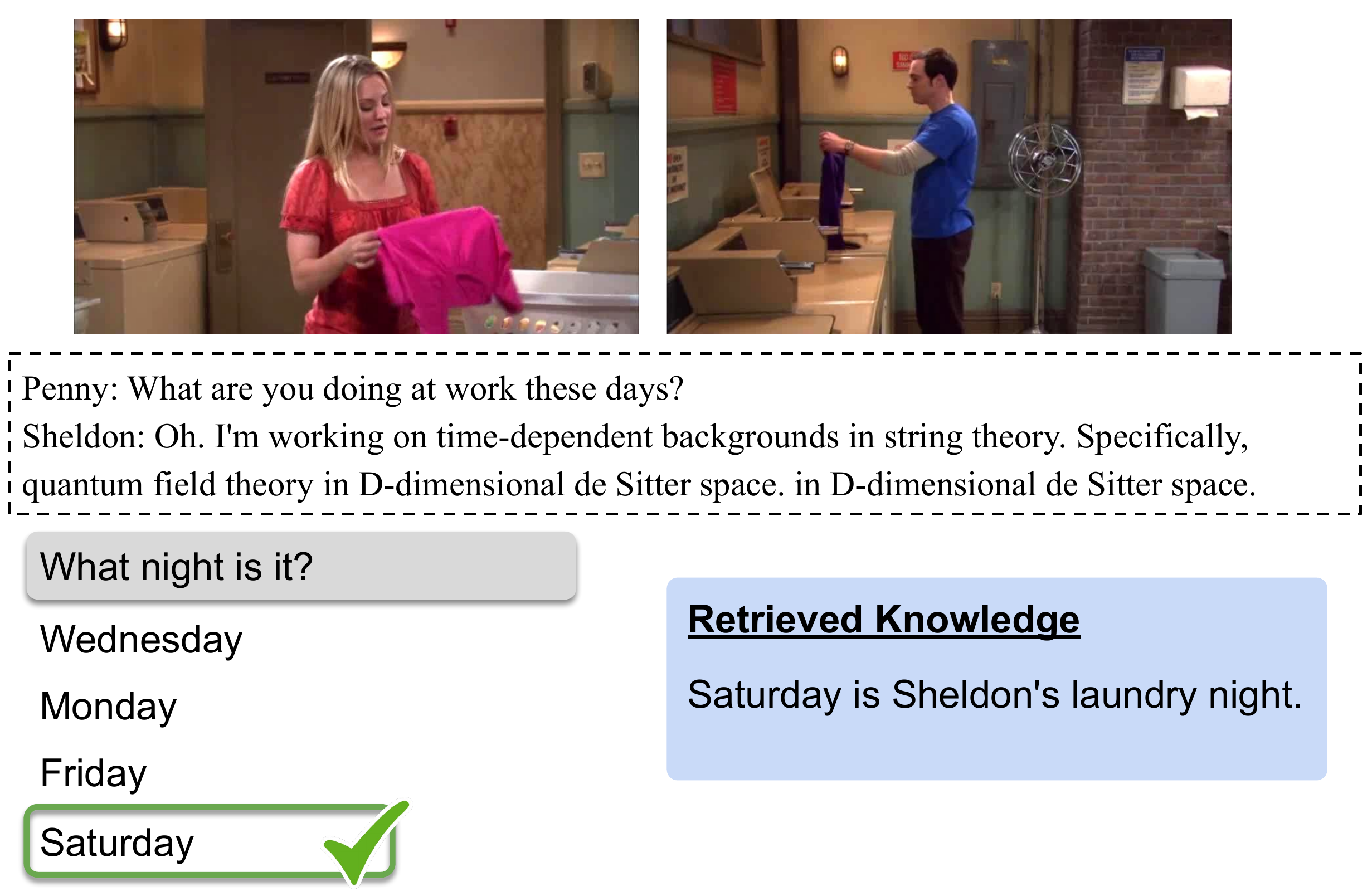}
\caption{KnowIT VQA sample correctly predicted by our model.} 
\label{fig:visualresults}
\end{figure}

This work contributes towards building a general framework in which different types of VQA coexist. To that end, we created the KnowIT VQA dataset with questions that requiere both video understanding and knowledge-based reasoning to be answered (example in Fig. \ref{fig:visualresults}). We use a popular sitcom as an ideal testbed for modelling knowledge-based questions about the world. We then cast the problem as a multi-choice challenge, and introduce a two-piece model that (i) acquires, processes, and maps specific knowledge into a continuous representation inferring the motivation behind each question, and (ii) fuses video and natural language content together with the acquired knowledge in a multi-modal fashion to predict the correct answer. The complete details about the dataset, model and exhaustive experimental results can be found in \cite{garcia2020knowit}.

%% file: 3-model.tex
\begin{figure*}[t]
\centering
\includegraphics[width = \textwidth]{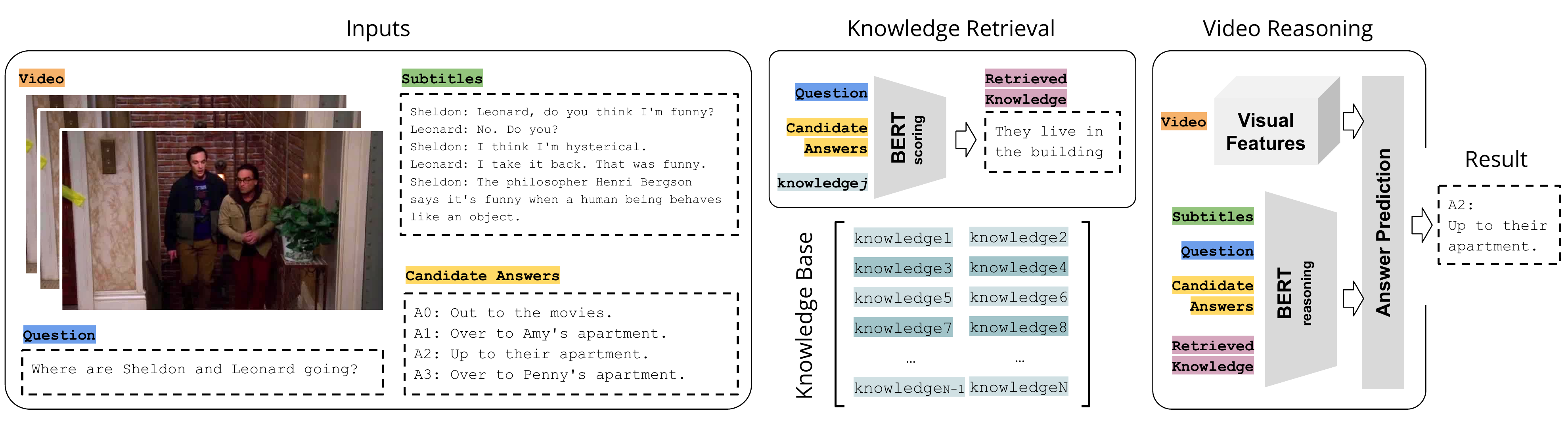}
\caption{Overview of our proposed model.} 
\label{fig:model}
\end{figure*}

\section{KnowIT VQA}
\label{sec:dataset}
The dataset is based on 207 episodes from the Big Bang Theory tv show. To generate questions, answers, and annotations we used Amazon Mechanical Turk.\footnote{https://www.mturk.com} We required workers to have a high knowledge about the show and instructed them to generate questions that are answerable by people familiar with the show, whereas difficult for new spectators. For each clip, we provided the video and subtitles, along with the summaries of all the episodes. Workers were asked to annotate each clip with a question, its correct answer, and three wrong but relevant answers. We randomly split the episodes into training, validation, and test sets, so that questions and clips from the same episode were assigned to the same set. In total, we gathered 24,282 question and answers pairs over 12,087 video clips.

In order to approximate the knowledge viewers acquire by watching the series, we annotated each video clip with expert information. We asked workers to describe in a short sentence the knowledge that is required to answer the question correctly. For example, for the question \textit{Why did Leonard invite Penny to lunch?}, the explanation \textit{Penny has just moved in} is key to respond the correct answer, \textit{He wanted Penny to feel welcomed into the building}. We called this field \textsc{knowledge}. We also annotated each question with four possible questions types: visual-, textual-, temporal-, and knowledge-based.

\section{ROCK Model}
\label{sec:model}

We propose ROCK (Fig.~\ref{fig:model}) a model for KBVQA in videos. ROCK consists of three different modules:

\textbf{1) Knowledge Base (KB)}: we build a specific KB, $ K = \{w_j\}$, with the \textsc{knowledge} field. Each instance, $w_j$ with {\small $j \in \{1,...,N\}$}, is represented as a natural language sentence.

\textbf{2) Knowledge Retrieval}:
We retrieve relevant information to each question in the KB. We input the question, the candidate answers, and a knowledge instance into a BERT network \cite{devlin2018bert}, namely BERT-scoring. BERT-scoring is trained to estimate a similarity score, $s_{ij}$, between a question, $q_i$, and a knowledge instance $w_j$,using matching (\ie $i = j$) and non-matching (\ie $i \neq j$) question-knowledge pairs and a binary cross-entropy loss. The top $k$ scoring knowledge instances for a given questions are retrieved.


\textbf{3) Video Reasoning}:
the visual and language content of the video clip is extracted and used to predict an answer.

\underline{Visual Representation}: We apply four different techniques to describe the visual content of video frames: 1) \textit{Image}, obtained from concatenating frame Resnet50 \cite{he2016deep} features; 2) \textit{Concepts}, bag-of-words with the objects and their attributes detected with \cite{Anderson2017up-down}; 3) \textit{Facial}, list of main characters in the clip detected with \cite{parkhi2015deep}; and 4) \textit{Captions}, sentences describing the visual content of the frames with \cite{xu2015show}.

\underline{Language Representation}: Textual data is processed using a fine-tuned BERT model, namely BERT-reasoning. We concatenate the captions, subtitles, question, a candidate answer, and the top $k$ retrieved knowledge instances and input it into the network. For each candidate answer, the output of the network is used as language representation.

\underline{Answer Prediction}: The visual and the language representations are concatenated and fed into a linear layer to obtain a candidate answer score. The answer with the highest score is the predicted answer. The network is trained as a classification task with the multi-class cross-entropy loss.

%% file: 4-evaluation.tex
\begin{table}
\caption{Accuracy for different methods on KnowIt VQA dataset. $\diamondsuit$ for parts of our model, $\bigstar$ for our full model.}
\centering
\resizebox{0.48\textwidth}{!}{\begin{tabular}{c l c c c c c }
\hline

& \textbf{Model} & \textbf{Vis.} & \textbf{Text.} & \textbf{Temp.} & \textbf{Know.} & \textbf{All}\\
\hline

\parbox[t]{3pt}{\multirow{6}{*}{\rotatebox[origin=c]{90}{Vis, Subs, QA}}}
& \texttt{TVQA}     & 0.612 & 0.645 & 0.547 & 0.466 & 0.522 \\
& $\diamondsuit$ \texttt{ROCK\textsubscript{VSQA}} {\scriptsize Image}              & 0.643 & 0.739 & 0.581 & 0.539 & 0.587 \\
& $\diamondsuit$ \texttt{ROCK\textsubscript{VSQA}} {\scriptsize Concepts}              & 0.647 & 0.743 & 0.581 & 0.538 & 0.587 \\
& $\diamondsuit$ \texttt{ROCK\textsubscript{VSQA}} {\scriptsize Facial}              & 0.649 & 0.743 & 0.581 & 0.537 & 0.587 \\
& $\diamondsuit$ \texttt{ROCK\textsubscript{VSQA}} {\scriptsize Caption}              & 0.666 & 0.772 & 0.581 & 0.514 &  0.580\\
& Humans {\scriptsize (Rookies)}           & 0.936 & 0.932 & 0.624 & 0.655 & 0.748 \\
\hline

\parbox[t]{3pt}{\multirow{5}{*}{\rotatebox[origin=c]{90}{Knowledge}}}
&  $\bigstar$ \texttt{ROCK} {\scriptsize Image}  & 0.654 & 0.681 & 0.628 & 0.647 & 0.652 \\
& $\bigstar$ \texttt{ROCK} {\scriptsize Concepts} & 0.654 & 0.685 & 0.628 & 0.646 & 0.652 \\
& $\bigstar$ \texttt{ROCK} {\scriptsize Facial} & 0.654 & 0.688 & 0.628 & 0.646 & 0.652 \\
& $\bigstar$ \texttt{ROCK} {\scriptsize Caption} & 0.647 & 0.678 & 0.593 & 0.643 & 0.646 \\
& Humans {\scriptsize (Masters)}             & 0.961 & 0.936 & 0.857 & 0.867 & 0.896 \\
\hline
\end{tabular}}
\label{tab:results}
\end{table}

\section{Evaluation}
\label{sec:evaluation}

We train our models with stochastic gradient descent with 0.9 momentum and 0.001 learning rate. For BERT, we used the uncased base model with pre-trained initialisation. We compare ROCK against two categories of models:

\textbf{1) Vis, Sub, QA.} Models using language and visual representations, but not knowledge, including (i) \texttt{TVQA} \cite{lei2018tvqa} a state-of-the-art VideoQA method in which language is encoded with a LSTM layer, whereas visual data is encoded into visual concepts; (ii) \texttt{ROCK\textsubscript{VSQA}} our model without knowledge; and (iii) Humans {\scriptsize (Rookies)} evaluators who have never watched any episode. Our model outperforms \texttt{TVQA} by 6.6\% but still lags well behind human accuracy

\textbf{2) Knowledge.} Models that exploit \textsc{knowledge} to predict the correct answer, i.e.~our \texttt{ROCK} model in its full version and Humans {\scriptsize (Rookies)} evaluators that have watched the show. Compared to the non-knowledge methods, the inclusion of the knowledge retrieval module increases the accuracy by 6.5\%, showing the great potential of knowledge-based approaches in our dataset. Among the visual representations, Image, Concepts, and Facial perform the same. However, the gap between ROCK and humans increases when knowledge is used, suggesting potential room for improvement in both video modeling and knowledge representation. An example result can be seen in Fig.~\ref{fig:visualresults}.

%% file: 5-conclusion.tex
\section{Conclusion}
\label{sec:conclusion}

We presented a novel dataset for knowledge-based visual question answering in videos and proposed a video reasoning model in which multi-modal video information was combined together with specific knowledge about the task. Our evaluation showed the great potential of knowledge-based models in video understanding problems. However, there is still a big gap with respect to human performance.

%% file: main.bbl
\begin{thebibliography}{10}\itemsep=-1pt

\bibitem{Anderson2017up-down}
Peter Anderson, Xiaodong He, Chris Buehler, Damien Teney, Mark Johnson, Stephen
  Gould, and Lei Zhang.
\newblock Bottom-up and top-down attention for image captioning and visual
  question answering.
\newblock In {\em CVPR}, 2018.

\bibitem{devlin2018bert}
Jacob Devlin, Ming-Wei Chang, Kenton Lee, and Kristina Toutanova.
\newblock {BERT}: {P}re-training of deep bidirectional transformers for
  language understanding.
\newblock In {\em NAACL}, 2019.

\bibitem{garcia2020knowit}
Noa Garcia, Mayu Otani, Chenhui Chu, and Yuta Nakashima.
\newblock Know{IT} {VQA}: Answering knowledge-based questions about videos.
\newblock In {\em AAAI}, 2020.

\bibitem{he2016deep}
Kaiming He, Xiangyu Zhang, Shaoqing Ren, and Jian Sun.
\newblock Deep residual learning for image recognition.
\newblock In {\em CVPR}, 2016.

\bibitem{lei2018tvqa}
Jie Lei, Licheng Yu, Mohit Bansal, and Tamara~L Berg.
\newblock {TVQA}: Localized, compositional video question answering.
\newblock In {\em EMNLP}, 2018.

\bibitem{parkhi2015deep}
Omkar~M Parkhi, Andrea Vedaldi, Andrew Zisserman, et~al.
\newblock Deep face recognition.
\newblock In {\em BMVC}, 2015.

\bibitem{tapaswi2016movieqa}
Makarand Tapaswi, Yukun Zhu, Rainer Stiefelhagen, Antonio Torralba, Raquel
  Urtasun, and Sanja Fidler.
\newblock Movieqa: Understanding stories in movies through question-answering.
\newblock In {\em Proceedings of the IEEE conference on computer vision and
  pattern recognition}, pages 4631--4640, 2016.

\bibitem{wang2018fvqa}
Peng Wang, Qi Wu, Chunhua Shen, Anthony Dick, and Anton van~den Hengel.
\newblock {FVQA}: {F}act-based visual question answering.
\newblock {\em TPAMI}, 40(10), 2018.

\bibitem{wu2016ask}
Qi Wu, Peng Wang, Chunhua Shen, Anthony Dick, and Anton Van Den~Hengel.
\newblock Ask me anything: Free-form visual question answering based on
  knowledge from external sources.
\newblock In {\em Proceedings of the IEEE Conference on Computer Vision and
  Pattern Recognition}, pages 4622--4630, 2016.

\bibitem{xu2015show}
Kelvin Xu, Jimmy Ba, Ryan Kiros, Kyunghyun Cho, Aaron Courville, Ruslan
  Salakhudinov, Rich Zemel, and Yoshua Bengio.
\newblock Show, attend and tell: {Neural} image caption generation with visual
  attention.
\newblock In {\em ICML}, 2015.

\end{thebibliography}
